\documentclass[letterpaper, 10 pt, conference]{ieeeconf}  % Comment this line out
\usepackage[english]{babel}
\usepackage[utf8]{inputenc}
\usepackage{algorithm,algpseudocode}
\usepackage{makecell}
\usepackage{pbox}
\usepackage{amsmath}
\usepackage{mathptmx}
\usepackage{graphicx}
\usepackage{multirow}
\usepackage{accents}
\usepackage{cite}
\usepackage{comment}
\usepackage{ragged2e}
\usepackage{bm}
\usepackage{url}
\usepackage{microtype}
\usepackage[font=scriptsize]{subcaption}
\usepackage{scalerel}
\usepackage{amsmath}
\usepackage{color}
\usepackage{listings}
\usepackage{float}
\usepackage{hyperref}
\usepackage{multicol}
\usepackage{acronym}
\acrodef{UPN}{Universal Planning Networks}
\usepackage{booktabs} % for professional tables
\usepackage{layouts}
\usepackage[autostyle]{csquotes}

 \let\oldReturn\Return
\renewcommand{\Return}{\State\oldReturn}                                            % if you need a4paper
\IEEEoverridecommandlockouts                              % This command is only
\allowdisplaybreaks[2]
\renewcommand{\baselinestretch}{0.95}
\title{\LARGE \bf
\vspace{-0.4cm} A Data-Efficient Framework for Training and Sim-to-Real Transfer of Navigation Policies \vspace{-0.2cm}
}

\author{Homanga Bharadhwaj$^{*1}$, Zihan Wang$^{*2}$, Yoshua Bengio$^3$ and Liam Paull$^3$ \vspace{-0.2cm}% <-this % stops a space
\thanks{*Equal contribution. Author ordering determined by Liam holding a duckie behind his back and making Homanga guess ``right'' or ``left''.}% <-this % stops a space
\thanks{$^1$ Homanga Bharadhwaj is with the Department of Computer Science and Engineering, IIT Kanpur, India}%
\thanks{$^2$ Zihan Wang is with the Division of Engineering Science, University of Toronto, Canada}%
\thanks{$^3$ Yoshua Bengio and Liam Paull are with Mila, Universite de Montreal}%
}
\begin{document}

\maketitle
\thispagestyle{empty} %UNCOMMENT AT THE END
\pagestyle{empty}
%\pagestyle{plain}

%%%%%%%%%%%%%%%%%%%%%%%%%%%%%%%%%%%%%%%%%%%%%%%%%%%%%%%%%%%%%%%%%%%%%%%%%%%%%%%%
\begin{abstract}
Learning effective visuomotor policies for robots purely from data is challenging, but also appealing since a learning-based system should not require manual tuning or calibration. In the case of a robot operating in a real environment the training process can be costly, time-consuming, and even dangerous since failures are common at the start of training. For this reason, it is desirable to be able to leverage \textit{simulation} and \textit{off-policy} data to the extent possible to train the robot. In this work,  we introduce a robust framework that plans in simulation and transfers well to the real environment. Our model incorporates a gradient-descent based planning module, which,  given the initial image and goal image, encodes the images to a lower dimensional latent state and plans a trajectory to reach the goal. The model,  consisting of the encoder and planner modules,  is trained through a meta-learning strategy in simulation first. We subsequently perform adversarial domain transfer on the encoder by using a bank of unlabelled but random images from the simulation and real environments to enable the encoder to map images from the real and simulated environments to a similarly distributed latent representation. By fine tuning the entire model (encoder + planner) with far fewer real world expert demonstrations, we show successful planning performances in different navigation tasks. 
\end{abstract}

%%%%%%%%%%%%%%%%%%%%%%%%%%%%%%%%%%%%%%%%%%%%%%%%%%%%%%%%%%%%%%%%%%%%%%%%%%%%%%%%
\section{INTRODUCTION}

Applying machine learning - and specifically deep reinforcement learning - to robotics algorithm development has shown great promise recently \cite{DRLRobotics1,DRLRobotics2,DLRobotics1}. 
However, state-of-the-art methods still require a lot of experiments on the physical robot \cite{Levine_IJRR_2018}, which is very expensive and possibly even dangerous if the robot is learning a task where wrong execution can cause harm or damage. Furthermore, there are few guarantees that a policy learned by one robot in a particular environment will \textit{transfer} to another (even slightly) different robot or another (even slightly) different environment.  
The recently popularized theory of ``meta-learning'' \cite{MAML,PMAML,BMAML} offers a methodology for overcoming the policy transfer issue, but at the expense of an even higher data requirement. 

In practice, a roboticist has two potential tools to aid in reducing the number of real on-policy rollouts that are needed on the real robot. The first is a \textit{simulator}. A simulator requires development effort to build, but there are now incredible tools to facilitate this. However, there will always be a discrepancy between the simulator and the real world, both in terms of the world dynamics and the perception of the environment. This will induce a distributional shift between training and test data which is problematic for deep learning. The second resource that is likely readily available is \textit{off-policy} rollouts from the real robot. The most common example could be data collected while the robot is being teleoperated safely by a person.

\begin{figure}
\includegraphics[angle=90,width=\columnwidth]{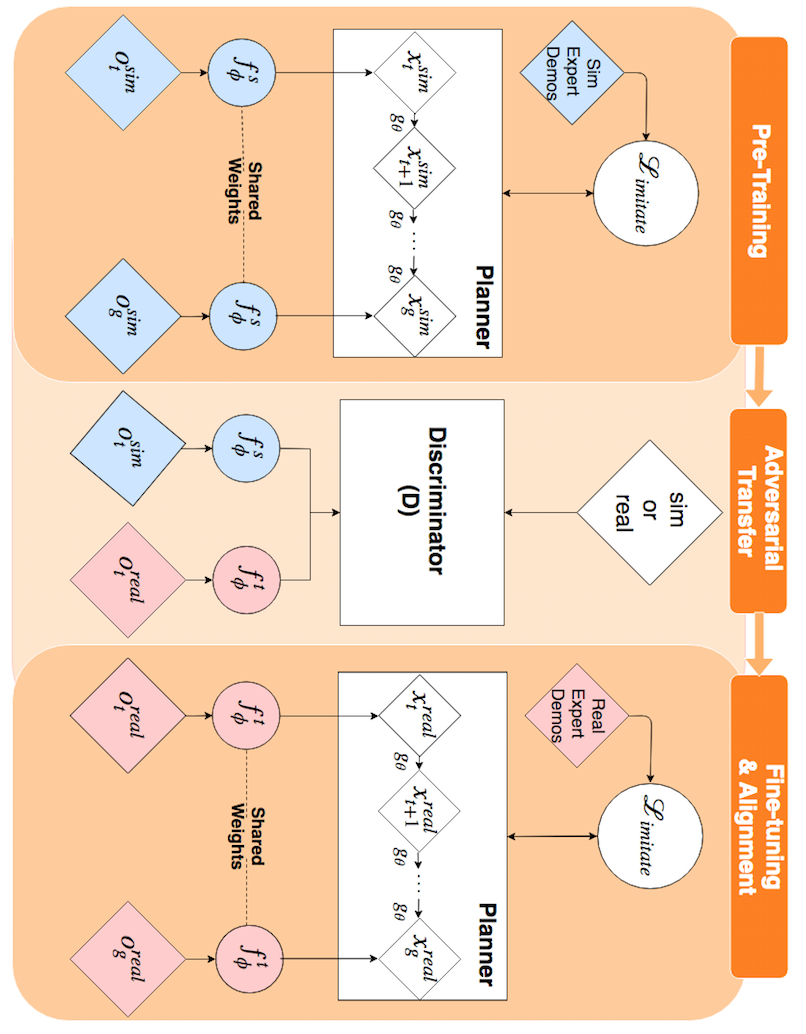}
\caption{\textbf{Sim-to-real transfer of navigation policies}: Our method comprises three main elements. First, a planner is trained in simulation. This process includes a learned image encoding $f_\phi^s$ and a learned dynamics model $g_\theta$. Second, an adversarial discriminative transfer approach is used to allow real images to be encoded in the same way as the simulated ones. Finally, a small amount of fine-tuning is performed on the real environment.}
\label{fig:flowchart}
\end{figure}

%Executing robotic learning in the simulator has the scope of swifter and more-scalable development of algorithms while leveraging lower-cost data collection. Especially for autonomous driving and navigation tasks, data collection in the real environment is not only difficult and costly but also very dangerous. Deep Reinforcement Learning [cite] and Imitation Learning [cite] algorithms have recently achieved human-level performance [cite]. 

%In the case of navigation, there are recent approaches [cite UPN VIN] that achieve state-of-the art results in simulation environments, while there are approaches of effective visuomotor policies trained and deployed  on the real environment [cite Levine / Abbeel papers, Levine's JMLR paper, some new followup papers]. Using learning based approaches for autonomous driving has not gained popularity because of the aforementioned difficulties of data collection and training in the real environment. 

In this work, we propose a novel procedure for combining these two resources (simulation and off-policy data) to efficiently train a physically embodied agent to complete a task in the real world. In short, we use the simulation environment to learn a policy for navigation in a meta-learning setup and then transfer the learned policy to the real world using an adversarial domain adaptation approach \cite{ADT}. 
%We introduce several theoretical
%\YB{(really theoretical?)} \HB{Prof Yoshua, we have introduced numerous modifications in the Planner (over the basic UPN structure) - Sec III B } \LP{I agree with Yoshua that theoretical might not be the right word here since these are real new theoretical ideas in the larger context of ML but are novel applications  } 
We use as a basis for our planner the Universal Planning Network \cite{UPN} but make several improvements that make our approach particularly well-suited to the transfer learning scenario and show the impact of these improvements by rigorous experiments on a real robot in the easily reproducible Duckietown environment~\cite{duckietown}.

Also of note is that most of the approaches in the literature related to the transfer from a simulation to a real robot that we are aware of consider a robot agent that is fully observed from an off-board camera. None of them consider the task of mobile robot navigation \cite{SimtoRealProgressive,dynamicsRand,ADT,SimtoRealHandEye,SimtoRealQuad}. 
In this work, we consider the case of a mobile robot with an onboard camera. This is an important consideration because the robot must now additionally implicitly infer its own state from partial observations over time rather than having the luxury to be able to infer its state fully from one observation. 
It is also more challenging from a visuomotor policy learning perspective since the camera itself is moving and therefore many of the pixels will change, rather than just the agent as in other works \cite{UPN}.

%This paper serves to bridge this gap by proposing a framework for training using a large amount data from simulation and very few trajectories from the real environment. Although there have been previous approaches of sim-to-real transfer of visuomotor policies [cite ADT- Corke, Domain Randomization of Abbeel], but to the best of our knowledge no previous work has explicitly tackled the problem of sim-to-real transfer for navigation tasks. 
% \LP{the following belongs in related work not here}\HB{Agreed} The paper [cite ADT] proposes an adversarial discriminative transfer approach for transfer of regression task and investigates transfer of visuo-motor policies from simulation to the real world. The explicit problem scenario considered by them is an object-reaching task wherein the model is clearly demarcated into perception and control modules, each trained separately using supervisory signals.

We also generalize the adversarial domain transfer method for sim-to-real transfer of an end-to-end gradient-descent based planner, where separate supervisory signals are not available for the perception and control modules separately. 
% This is typically the case in navigation as opposed to a simple object-reaching task. Our model embeds a differentiable planner that plans a sequence of actions from the initial image to the goal image. The images are first encoded in the latent space (perception module) and then by repeated gradient-descent based updates, a sequence of actions are optimized in latent space. The only supervisory signals provided are from expert trajectories used to update the parameters of the planner and the encoder.
We first train using expert trajectories in simulation and then perform adversarial transfer on the encoder's output space to learn mappings from the real environment that are similar to the mappings from the simulation environment. 
% To facilitate the adversarial transfer, we use a few real trajectories as well to provide supervisory signals. 
In particular, we claim the following contributions: 
\begin{itemize}
    \item We develop a stable and efficient planning model for navigation  through incorporation of a meta-learned loss function, latent space regularization terms and a stochastic forward dynamics model in the planning objective. 
    \item We demonstrate on a real robot that the developed policy (encoder + planner)  trained in simulation can transfer to a real environment (by using very few real expert demonstrations for fine-tuning) through an adversarial transfer approach. 
\end{itemize}

\section{BACKGROUND AND RELATED WORKS}

Our work draws inspiration from recent developments in meta-learning and sim-to-real policy transfer.

\subsection{Meta Learning}

Meta-learning models are trained by being subjected to a variety of tasks in training and are then tested in their ability to learn new tasks. The concept is not new~\cite{MetaBengios,MetaSchmidhuber}, but has become increasingly relevant in modern deep reinforcement learning and imitation learning algorithms~\cite{MetaNew1, MetaNew2,MAML,MetaVictor, DBLP:journals/corr/SantoroBBWL16, grant2018recasting,DBLP:journals/corr/abs-1802-04821, sprechmann2018memorybased}.  Model-Agnostic Meta Learning (MAML)~\cite{MAML,PMAML,BMAML} provides a framework for rapidly adapting gradient-based planners to different (new) tasks by performing a few gradient steps. On a high level, our approach is inspired by MAML in the sense that we have a two-stage computation through gradient descent during training. The inner stage computes a plan given the planner, while the outer stage updates the parameters of the planner, including the weights of the neural network used as the inner stage loss function. 
%A similar strategy was used in UPN~\cite{UPN}, where the outer loop updates the parameters of the inner loop planner by imitation learning.
% \HB{Prof Liam, do we need more meat here?} \LP{we could use a bit more of a survey here}

\subsubsection{Universal Planning Networks}

The UPN~\cite{UPN} framework considers the problem of finding a plan $\hat{a}_{t:t+T}$ given an initial image $o_{t}$ and a goal image $o_{g}$ as inputs. Similar to MAML it employs a two-tiered approach: 1) optimize the trajectories (sequence of actions) with gradient decent given a planner (inner loop) and 2) optimize the representations in the planner (outer loop) using expert trajectories. The planning module consists of a forward dynamics model $g_{\theta}$ (a fully connected neural network) and an encoder $f_{\phi}$ (a convolutional neural network) with $\theta$ and $\phi$ being neural network parameters respectively, which are learned in an end-to-end manner.

In each iteration, for a fixed planning horizon $T$, the current and goal images are encoded into a latent space $\mathcal{X}$:
\begin{equation}
    \begin{split}
        x_{t} &= f_\phi(o_{t})\\
        x_{g} &= f_\phi(o_{g})
    \end{split}
\end{equation}
The latent representation at the end of the horizon, $x_{t + T + 1}$ is calculated by recursively applying the learned forward dynamics model and the current estimate of the actions, $\hat{a}_{t...t+T}$ in the planned trajectory:
\begin{equation}
\hat{x}_{t+1} = g_\theta(\hat{x}_{t}, \hat{a}_{t})
\end{equation}
starting from the latent encoding of the initial image $x_{t}$.
The inner loop planning loss is then calculated as the discrepancy between the direct encoding of the goal image and the latent space estimate generated by propagating the initial image encoding through the learned dynamics model $T$ times.
\begin{equation}
    \mathcal{L}_{plan}^{(i)} =  ||\hat{x}_{t+T+1}^{(i)} -  x_{g}||^2 
\end{equation}
This loss is backpropagated to find the best \emph{actions} given the encoding parameters $\phi$ and the dynamics model parameters $\theta$. This process repeats until convergence (gradient descent).
%The actions are updated repeatedly by back-propagating the inner loop planning loss  for a fixed number of iterations corresponding to particular values of $\theta$ and $\phi$. 
Once a trajectory has been converged upon, it is compared with an expert trajectory, $a^{*}_{t:t + T}$,  using an outer-loop imitation loss:
\begin{equation}
    \mathcal{L}_{imitate} = ||\hat{a}_{t:t + T} - a^{*}_{t:t + T}||^2_2
\end{equation}
This loss is back-propagated into the planner and used to update the parameters of the planner $\phi$ and $\theta$. This process continues over a batch of expert demonstrations until convergence in the hope that the resulting latent space encoding and dynamics model parameters will be automatically learned.

This setup is elegant since it is able to learn a latent encoding without  wasting additional optimization effort on reconstruction as is the case in a variational autoencoder setup such as DARLA \cite{DARLA}. However, in our experience it suffers from the following shortcomings:
\begin{enumerate}
    \item It is data inefficient and requires a lot of expert trajectories to train,
    \item The inflexible planning loss constrains the learning process because it is not necessarily suitable for every task, since what is a good representation to model state transitions may not be best to measure discrepancy to the goal
    \item While it is able to adapt to new dynamics models (this is shown in an RL context in \cite{UPN}) it is not able to adapt to changes in the perceptual environment, which limits its ability to transfer from a simulator to a real robot,
    \item The learned dynamics model lacks the robustness to be used on a real robot since it is devoid of any notion of stochasticity. 
\end{enumerate}

In Sec.~\ref{sec:method} we detail how our method overcomes these shortcomings.

\subsection{Sim-to-Real Transfer}

The goal of sim-to-real transfer is to use simulated or synthetic data, which are cheap and easy to be collected, to partially or fully replace the use of real-world data, which are expensive and time consuming to be obtained~\cite{Sim2RealJamesDJ17, Sim2RealSadeghiL16, DomainRand}. The main challenge in effective sim-to-real transfer is that there are aspects of reality which cannot be modelled well in the simulation environment~\cite{SimtoReal-whysimfail}. Hence, a model that has been trained in simulation cannot be directly deployed in the real environment since there is a distributional shift between the test data and the training data~\cite{SimZhangLMUC15}. One approach to close the ``reality gap" is by matching the simulator to physical reality via dedicated system identification and superior-quality rendering~\cite{SimtoReal-PhyEng1,SimtoReal-PhyEng2, ViereckPSP17}. However this is very expensive in terms of development effort and, not very effective based on past results~\cite{RGBRendering}. Apart from this, there are broadly two categories of approaches to resolve the aforementioned issue, 1) learning  invariant features  and 2) learning a mapping from simulation to real.    \\

\subsubsection{Learning Invariant Representations}

Domain randomization~\cite{DomainRand,Sim2RealJamesDJ17, dynamicsRand,DomainRand1, DomainRandLerrel, Tzeng2015TowardsAD, Sim2RealSadeghiL16} bridges the reality gap by leveraging rich variations of the simulation environment during training. The hope is that by adding random variability in the simulator, the real data distribution will be within that of the training data.  
%The simulator is randomized and the model is exposed to a wide range of environments during training with the hope that this will enable learning a robust policy that can directly adapt to the intricacies of the real world. 
However, recent results have only been able to successfully use domain randomization for relatively simple tasks like object localization~\cite{DomainRand} and robotic grasping~\cite{DomainRandGrasp} with no use cases in navigation to the best of our knowledge. Additionally, which parameters to randomize and to what degree is done heuristically and requires significant testing and tuning.  \\

\subsubsection{Learning the Mapping between Simulation and Real}
%We choose the second\\
A second option is to explicitly learn the relationship between the simulated and real data~\cite{Sim2RealZhangLUC16}. Then, a policy trained on the simulator can be executed in the real world by pre-processing the real data to make it seem like simulated data. 
A recent approach~\cite{DBLP:journals/corr/ShrivastavaPTSW16} proposed a Simulated+Unsupervised (S+U) learning method which utilizes unlabeled real data to learn a model in order to improve the performance of a simulated agent. A Generative Adversarial Network (GAN) was trained to distinguish the nature of the images (sim or real) and improve the quality of the image encoder.

Another approach, namely ``Adversarial Discriminative Domain Adaptation" \cite{ADTPrecursor} has the key advantage over prior methods of not requiring pair-wise labeled data from the two domains. 
%This data can be very challenging or even impossible to obtain. Rather, in the adversarial discriminative setting, 
All that is required is batches of data from each domain and labels corresponding to their ground truth domain. The GAN approach builds a representation that attempts to fool a discriminator as to the true origin of the data thereby learning a mapping from one domain to the other. 

This was recently applied to sim-to-real transfer for a robotic table-top-reaching task with a 7 DoF arm \cite{ADT}. The authors show the ability to  effectively transfer the learning of a visuomotor policies from a simulation environment to the real setup by the use of very few real expert demonstrations for fine-tuning. The architecture consists of two key components:
\begin{itemize}
    \item A perception module  that estimates the object position $\mathbf{x^*}$ from a raw-pixel image $I$ (based on a VGG16 neural network \cite{VGG});
    \item A control module that estimates the optimum joint velocities $\mathbf{v}$ given the position $\mathbf{x^*}$ and joint angles $\mathbf{q}$.
\end{itemize} 

The source encoder is first pre-trained using labelled simulated data of images and corresponding target positions. Then, the source encoder ($E_s$) is locked and a reference target  encoder ($E_r$) is trained through images sampled from both the simulation ($I^s$) and the real ($I^r$) setup. They use an adversarial loss 
    $L_{Ad} = L_D + \gamma L_E$
where 
\begin{equation}
    \begin{split}
        L_D &= -\frac{1}{2m}\sum_j [ \log D(E_s(I_j^s)) + \log (1- D(E_r(I_j^r)))] \\
        L_E &= -\frac{1}{m}\sum_j \log D(E_r(I_j^r))
    \end{split}
\end{equation}  
Here, $D$ denotes the discriminator and $\gamma$ is a balancing weight. In practice the authors use a supervised loss over real expert demonstrations in addition to the adversarial loss for successful transfer.
% Similarly, another recent approach uses an adversarial discriminative method for the transfer of a perception module from simulated to real environments ~\cite{ADT}. 
This method is appealing since it provides a principled way to transfer learned policies from simulation to the real robot with limited and not necessarily pairwise matched labeled data from the real robot. However, the authors explicitly consider the output of the perception module to correspond to object position and formulate the control module to map from positions to velocities. Letting the image encoding of the perception module correspond to position restricts the wide scope of latent features that can be encoded, and hence we do not explicitly force the encoding in our model to correspond to one particular tangible attribute (like position). However this introduces a difficulty in sim-to-real transfer because there is no ground-truth supervision for the perception module alone. In our proposed method, we train end-to-end in simulation and hence require no ground truth perceptual data, only a select number of expert trajectories to be used in the outer-loop imitation learning loss.

\vspace*{-0.1cm}
\section{METHOD}
\label{sec:method}
The basis of our approach is inspired from two areas of recent rapid development: meta-learning for planning, and discriminative policy transfer. 
An overview of the approach is shown in Fig.~\ref{fig:flowchart}.
\vspace*{-0.2cm}

\subsection{Proposed End-to-End Planner}

% Given an initial image and a goal image, the planner first encodes the images to a latent state which serves as a proxy for the unobserved state of the agent. 
We build our planner, which consists of the encoder $f_{\phi}$, the forward dynamics model $g_{\theta}$ and the planning loss $\mathcal{L}_{plan}$ in a UPN-style framework. 
%In what follows we describe how we overcome the limitations of UPN detailed in Sec.~\ref{sec}.

\subsubsection{Stochastic Forward Dynamics Model}

In UPN~\cite{UPN}, the forward dynamics model $g_{\theta}$ is fully deterministic, which  makes the model inappropriate when applied to a real robot, since transitions are not deterministic (and especially if the next state conditioned on the previous state is not unimodal), as well as making the model brittle to slight perturbations in the initial and/or goal image. 
We capture this intuition for making our model robust by explicitly encoding \textit{noise} in the dynamics model:
\begin{equation}
    \hat{x}_{t+1} \sim g_{\theta}(\hat{x}_t,\hat{a}_t,\epsilon)
    \label{eq:dynamics}
\end{equation}
where $\epsilon$ is sampled from a zero-mean, fixed variance normal distribution. %\HB{I think instead of equality we should replace it by a sample (tilde) $\sim$}

% Hence, in every iteration of forward computation, the predicted latent state $\hat{x}_{t}$ at each time-step is a sample from a normal distribution. 
\subsubsection{Learning the Planning Loss Function}
\label{sec:MLP}
Most existing approaches~\cite{UPN,VIN,MAML,DomainRandGrasp} use a fixed loss function, like squared error loss or Huber loss~\cite{UPN}. We alleviate the modelling bias introduced by a fixed loss function by adopting one with tunable parameters. In particular, we use a Multi-Layer Perceptron (MLP) as the planning loss, the parameters of which are ``meta-learned'' through the outer loop imitation loss. Our new inner loop planning loss becomes:
\begin{equation}
    \mathcal{L}_{MLP} = MLP (\hat{x}_{g}, x_{g})
\end{equation} 
The  intuition behind using an MLP as the loss function is to let the model suitably adapt the loss function to any particular task by tuning the parameters of the MLP. 

% A natural question is whether this method will require a lot more training trajectories for achieving optimum performance because an additional NN needs to be trained. We show through detailed simulation studies that the costs incurred are negligible in comparison to the performance boost achieved.  

\subsubsection{Faster Convergence through Regularization}
The original UPN framework is relatively data inefficient since all information about the latent encoding parameters and the dynamics model must be learned from the outer loop imitation loss. 
We propose two forms of regularization to the model to alleviate this.

The first is a ``smoothness" regularization which enforces the successive latent states to be ``close'' to each other in latent space. 
%The latent state at time $t$, which we denote by $x_t$ is expected to encode the complete information of the state of the agent at that time. 
Since, the transition from $\hat{x}_{t}$ to $\hat{x}_{t+1}$ occurs as a result of action $\hat{a}_t$ on a physical robot  (i.e., $\hat{x}_{t+1} \sim g_{\theta}(\hat{x}_t,\hat{a}_t,\epsilon)$) we should expect that, in order to have a smooth trajectory, the ``distance'' in latent space between subsequent state encodings should be small. 
%$||\hat{x}_t -  \hat{x}_{t+1}||_p$ to be small ($\forall t$), where $||\cdot||_p$ denotes the $L_p$ norm. 
We enforce this by adding the the following term to the planning loss: 
\begin{equation}
    \mathcal{L}_{smooth} = \sum_{t=t}^{t=g}||\hat{x}_t -  \hat{x}_{t+1}||_p
    \label{eq:loss-smooth}
\end{equation}
where $||\cdot||_p$ denotes the $L_p$ norm. Note that since $g_{\theta}(\hat{x}_t,\hat{a}_t,\epsilon)$ is a distribution, $\hat{x}_t$ is a sample from that distribution. 

The second type of regularization enforces ``consistency''.
The original planning loss enforces a notion of consistency but only at the terminal state $x_{g}$. By consistency, we mean that the error represents the discrepancy between the terminal latent states calculated two ways: 1) by encoding the goal image and 2) by encoding the initial image and propagating the latent state through the dynamics model $T$ times.
However, in practice during training we have the entire sequence of images. Therefore, we can enforce consistency at each timestep \textit{regardless of the policy being executed to generate the data}. This is achieved by considering the two pathways that we can use to arrive at the same latent state: 1) encode image at time $t$ and propagate through the dynamics model and 2) encode the image at time $t+1$
%incorporates the notion of how well is the re-construction of the inferred final latent state as compared to the encoded final latent state. We argue that for successful planning, this reconstruction must be reasonably good at each planning step. 
More precisely, we enforce that $g_{\theta}(f_{\phi}(o_t),a_t,\epsilon)$ and $f_{\phi}(o_{t+1},\epsilon)$ are ``close'' to each other in distribution at every time-step $t$ by adding:
\begin{equation}
    \mathcal{L}_{consist} = \sum_{t=t}^{t=g}||g_{\theta}(f_{\phi}(o_t),a_t,\epsilon)-f_{\phi}(o_{t+1},\epsilon)||_p
    \label{eq:loss-consistency}
\end{equation}
to the planner loss function. Here, the two terms are samples from the respective distributions in each rollout. Note that here, $a_t$ is sampled to be either the expert action (with a probability of 80\%) or the current action (being optimized) at time-step $t$ and $o_{t+1}$ is the observed image at time-step $t+1$ after the agent takes action $a_t$ in the state with observation $o_t$. An overview of the training process is outlined in Alg.~\ref{alg:pre-training}.
%So, finally $\mathcal{L}_{plan} = \mathcal{L}_{MLP} + \mathcal{L}_{smooth} + \mathcal{L}_{consistency}$.
%For consistency regularization we incorporate the term $\sum_{t=t}^{t=g}||g_{\theta}(f_{\phi}(o_t),a_t,\epsilon)-f_{\phi}(o_{t+1},\epsilon)||_p$ in the planning loss.

% Here, $a_t$ is either an expert action at time-step $t$ ($a^*_t$) or the current action being optimized by the planner ($\hat{a}_t$). We choose the expert action in our implementation. 

\subsection{Policy Transfer to the Real Robot}

{\small
\begin{algorithm}[t]
    \caption{Sim-to-Real Transfer of Navigation Policy}
\begin{algorithmic}

%\State /* Pre-training */
\State Randomly initialize $\theta,\phi,\zeta$
 \State $f_{\phi}^s$, $g_{\theta}^s$, $MLP_{\zeta}^s$ =
\Call{Training}{$a^{sim*}_{t:t+T}$, $\beta_1$, $\beta_2$, $\beta_3$, $\alpha$, $n_p$}
%\State /* Adversarial Transfer */
 \State $f_{\phi}^t$ = \Call{Transfer}{$\{o^{real} \}$, $\{o^{sim} \}$, $f_{\phi}^s$, $k$}
%\State /* Fine- Tuning  */
% \State Fine-tuning proceeds exactly as pre-training but instead of randomly initializing the weights of $f_{\phi}^t$, $g_{\theta}^s$ and $MLP_{\zeta}^s$, we use the above
\State $\phi\xleftarrow[]{}f_{\phi}^t,\theta\xleftarrow[]{}g_{\theta}^s,\zeta\xleftarrow[]{}MLP_{\zeta}^s$
 \State $f_{\phi}$, $g_{\theta}$, $MLP_{\zeta}$ = \Call{Training}{$a^{real*}_{t:t+T}$, $\beta_1$, $\beta_2$, $\beta_3$, $\alpha$, $n_p$}

\end{algorithmic}
\label{alg:overview}
\end{algorithm}
}
{\small
\begin{algorithm}[t]
    \caption{Planner Training}
\begin{algorithmic}

%\Function{Training}{$a^{*}_{t:t+T}$,$\beta_1$,$\beta_2$,$\beta_3$}
%\Require Expert demonstrations in sim $a^{*}_{t:t+T}$
\Procedure{Training}{$a^{*}_{t:t+T}$, $\beta_1$, $\beta_2$, $\beta_3$, $\alpha$, $n_p$}
\For{number of training iterations}
    \State Sample a batch of demonstrations $o_t, o_g, a^*_{t:g}$
\State Take Randomized guess for the optimal plan $\hat{a}_{t:g}^{(0)}$
\For{$i$ from $0$ to $n_p-1$}
    \State Compute $x_{t} = f_\phi(o_{t})$, $x_{g} = f_\phi(o_{g})$
    \For{$j$ from $0$ to $T$}
     \State       $\hat{x}_{t+j+1}^{(i)} = g_\theta(\hat{x}_{t+j}^{(i)},{\hat{a}_{t+j}}^{(i)}, \epsilon) $
    \EndFor
    \State Compute: $\mathcal{L}_{plan} = \mathcal{L}_{MLP} + \mathcal{L}_{smooth} + \mathcal{L}_{consist}$ 
    \State Update: ${\hat{a}_{t:t+T}}^{(i+1)} = {\hat{a}_{t:t+T}}^{(i)} - \alpha \nabla_{{\hat{a}_{t:t+T}}^{(i)}} \mathcal{L}_{plan}^{(i)} $
\EndFor
    \State Compute $\mathcal{L}_{imitate} = ||\hat{a}_{t:g} - a^{*}_{t:g}||^2_2$
    \State Update  $\theta := \theta - \beta_1 \nabla_{\theta}\mathcal{L}_{imitate}$
    \State Update $\phi := \phi - \beta_2 \nabla_{\phi}\mathcal{L}_{imitate}$
    \State Update $\zeta := \zeta - \beta_3 \nabla_{\phi}\mathcal{L}_{imitate}$
\EndFor
\Return $f_{\phi}$, $g_{\theta}$, $MLP_{\zeta}$
\EndProcedure
%\EndFunction
\end{algorithmic}
\label{alg:pre-training}
\end{algorithm}
}
{\small
\begin{algorithm}[t]
    \caption{Sim-to-Real Transfer}
\begin{algorithmic}
\Procedure{Transfer}{$\{o^{real} \}$, $\{o^{sim} \}$, $f_{\phi}^s$, $k$}
%\Function{Transfer}{$\{o^{real} \}$,$\{o^{sim} \}$,$f_{\phi}^s$,$k$}
%\Require Labelled Images from real $\{o^{real} \}$ and sim $\{o^{sim} \}$
%\Require $f_{\phi}^s$, which is the encoder of the pre-trained model 
\For{number of training iterations}
\For{$k$ steps}
\State Sample a batch of $N$ real images $o^{real}_{1:N}$ 
\State Sample a batch of $N$ sim images $o^{sim}_{1:N}$ 
\State Update Discriminator D : $\nabla_{\theta_d} L_D$
\EndFor
\State Sample a batch of $N$ real images $o^{real}_{1;N}$ 
\State Update Generator (Target Encoder) $f_{\phi}^t$ by ascending \State its stochastic gradient: $\nabla_{\theta_d} L_G$
\EndFor
\Return $f_{\phi}^t$ // Target encoder
\EndProcedure
%\EndFunction

% \State  /* Fine-tuning in real */

% \Require Target encoder $f_{\phi}^t$
% \Require Pre-trained forward dynamics model $g_{\theta}^s$
% \Require Pre-trained planning loss $MLP_{\zeta}^s$
% \Require (A few) Real Expert demonstrations $a^{real*}_{t:t+T}$
% \State Fine-tuning proceeds exactly as pre-training but instead of randomly initializing the weights of $f_{\phi}^t$, $g_{\theta}^s$ and $MLP_{\zeta}^s$, we use the above

\end{algorithmic}
\label{alg:transfer}
\end{algorithm}
}

Although a gradient-descent based planning algorithm is very general and powerful in the sense that it can be applied to different tasks, training through imitation learning is data intensive and requires many  demonstrations, something which is not always possible to collect in a real environment. Hence, training in simulation and fine-tuning in the real setup is a promising direction for using such architectures in real robotic tasks like navigation and grasping. However, it is not immediately evident if a sim-to-real transfer architecture can be applied in this framework because the latent encoding does not have an easily interpretable physical meaning. 

We propose a method based on pre-training in simulation, using an adversarial discriminative approach for policy transfer, followed by a fine-tuning approach on the real robot as detailed in Alg.~\ref{alg:overview}.

\subsubsection{Pre-training in simulation}
Expert trajectories are very inexpensive to obtain in a simulation (once the simulator has been built) and therefore this represents the bulk of our training phase. Further details are presented in Alg.~\ref{alg:pre-training}
%This phase basically entails using a lot of expert trajectories in simulation and the corresponding initial and goal images for training the architecture end-to-end. The training proceeds as described in Algorithm 1. 
%The $\mathcal{L}_{plan}^{(i)}$ loss of the planner is defined as follows 
%\begin{align*}
%     \mathcal{L}_{plan}^{(i)} = MLP_{\zeta} (\hat{x}_{g}, x_{g}) + \sum_{t=t}^{t=g}||g_{\theta}(f_{\phi}(o_t),a_t,\epsilon)-f_{\phi}(o_{t+1},\epsilon)||_p \\+ \sum_{t=t}^{t=g}||\hat{x}_t -  \hat{x}_{t+1}||_p 
%\end{align*}
% It is crucial that the agent trains well in the simulation for effective sim-to-real transfer.

\subsubsection{Adversarial transfer of encoder from sim-to-real}
Once we have a policy that is performing well in the simulator, we aim to learn an encoder that generates the same distribution of latent states over real images as the pre-trained encoder. To achieve  this we begin by freezing the  source encoder's learned weights. We feed in images sampled randomly from the simulation environment and execute one forward pass through the source encoder to yield a latent embedding 
    $x_{sim} = f_{\phi}^s(o^{sim})$
where $f_\phi^s(\cdot)$ is the simulator encoder.
We initialize the target encoder with the same weights as the source encoder but do not freeze them  (i.e. the weights of the target encoder are trainable). The target encoder is fed images randomly sampled from the real environment and we execute one forward pass to yield a latent embedding
    $x_{real} = f_{\phi}^t(o^{real})$
where $f_{\phi}^t(\cdot)$ is the real robot encoder. 

We then use a three-layer feedforward neural network as a discriminator ($D$) to distinguish between which latent representations are obtained from images of simulation and which are obtained from real images. This is an adversarial learning framework where the generator is the target encoder that tries to generate latent representations from real images which are close to the representations of the trained source encoder on images from simulation. The discriminator and generator losses used in Alg. \ref{alg:transfer} are:
\begin{equation}
\begin{split}
L_D &= -\frac{1}{2N}\sum_{i=1}^{N}[\log D(f_{\phi}^s(o^{sim}_i) + \log (1 -D(f_{\phi}^t(o^{real}_i)  ) ] \\
L_G &= -\frac{1}{N}\sum_{i=1}^{N}[\log D(f_{\phi}^t(o^{real}_i)  ]
\end{split}
\nonumber
\end{equation}
If the process of adversarial domain transfer is perfect, then without changing the rest of the architecture, the forward dynamics model $g_{\theta}^s$ and MLP loss function $MLP_{\zeta}^s$ pre-trained on simulation affixed to the target encoder $f_{\phi}^t$ should be able to perform well in the real environment. In practice, due to imperfect convergence of adversarial training, we need to incorporate fine-tuning with some expert demonstrations from the real environment. This is exactly similar to the pre-training phase, except for the fact that expert trajectories are from the real environment.

\section{EXPERIMENT DESIGN}

To test the performance of our architecture, we designed two experiments on the Duckietown~\cite{duckietown} platform: lane following and left turn. For each test run, we selected different initial poses for the Duckiebot, with each pose being a pair of initial position and initial facing angle. 

% In the duckietown environment, the track has been simplified into a two-lane road with normal broken yellow line in the middle and solid white lines on the sides, which followed the Ontario pavement marking scheme \textbf{[https://www.ontario.ca/document/official-mto-drivers-handbook/pavement-markings]}. The duckiebot was tested on the duckietown simulation first. After verifying the validity on simulator, then we implemented the algorithm on the real duckiebot to evaluate the performance in the real environment. 

% In order to test the generality, we have selected several initial poses for the duckiebot. The initial pose was composed of two parts: initial position and initial facing angle. 
In simulation, for the lane following test, we select the initial angles from the range -30º to 30º and the initial positions from the center of the right lane to the center of the left lane. For the left turn test, the initial angle ranges from -30º to 30º and the initial position ranges from the center of the right lane to the broken yellow (middle) line. We randomly generate a number of initial poses in the above mentioned ranges during testing and a number of expert trajectories of different horizon lengths during training. 
%For both training and testing, the goal image is fixed and corresponds to an image of the center of the road at an appropriate horizon.

\begin{figure}
\centering
\includegraphics[width=0.8\columnwidth,scale=0.5]{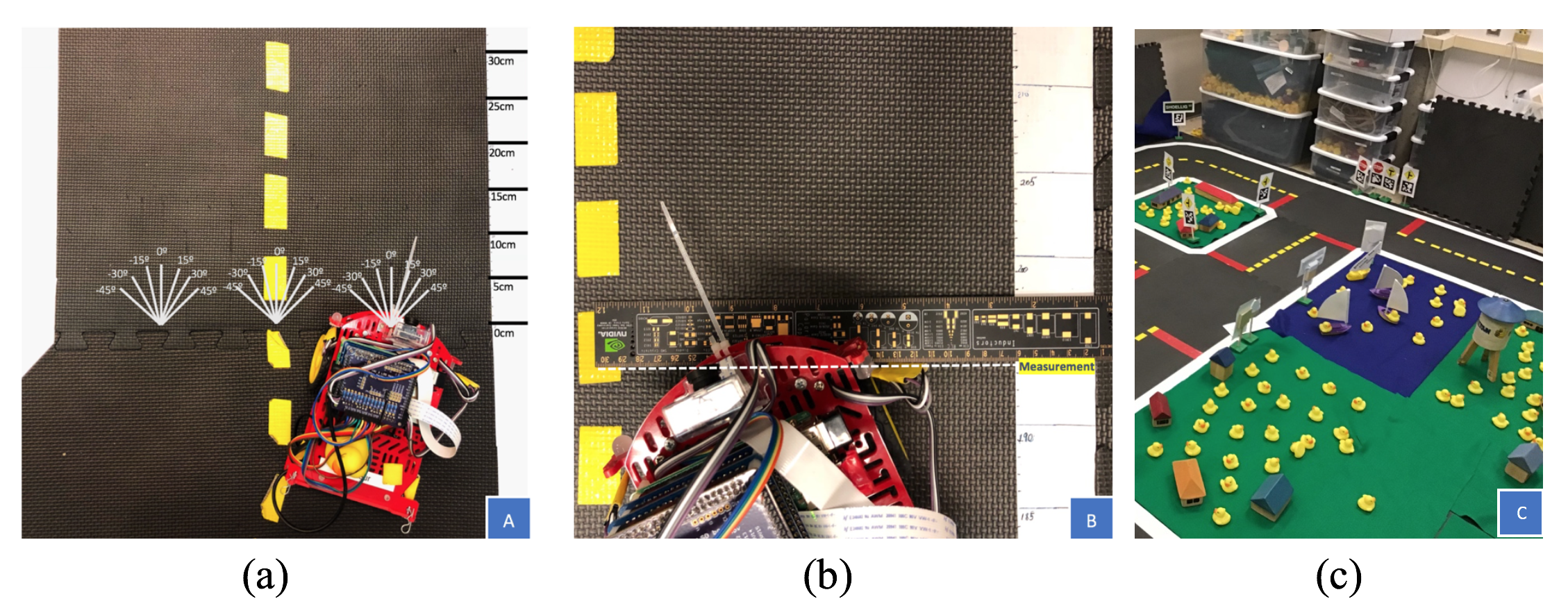}
\caption{\textbf{The Duckietown Environment}: (a) The initial pose setup for the Duckiebot in the real Duckietown environment. (b) A demonstration of distance measurement.  (c) An overview of the Duckietown environment. }
\label{fig:real_setup}
\end{figure}

In the real environment we uniformly discretize the space of initial poses. For lane following, there are three initial positions, namely center of the right lane, left lane and yellow line and seven values of initial angles (-45º, -30, -15º, 0º, 15º, 30º, 45º). For the left turn test, there is one intiial position, namely the center of the right lane and five initial angles (-30, -15º, 0º, 15º, 30º). See Figure~\ref{fig:real_setup}.

% In the simulation, in order to increase the diversity of the expert trajectories, the initial positions and angles were randomly selected from a range of values. For the lane following test, the initial positions were range from the center of right lane to the center of left lane. The initial angles were range from -45º to 45º. For the left turn test, we assume the duckiebot were driving properly on the right lane. The initial position range is from the right lane center to the center of the broken yellow line. The initial angles are range from -30º to 30º. In order to simplify the testing complexity in the real world, the initial positions for lane following were reduced to three specific values: left (the center of the left lane), middle (the center of the broken yellow line) and right (the center of the right lane) and the initial angles were reduced to seven specific values: -45º, -30, -15º, 0º, 15º, 30º, 45º. Similarly, the initial poses for the left turn were simplified to one initial position (right) and five initial angles (-30, -15º, 0º, 15º, 30º). 

\subsection{Dataset Collection}
The dataset for training consists of expert trajectories in simulation, expert trajectories in the real setup and images from both the simulator and real setup (sim/real frame data) in any context. The expert trajectories in both sim and real are collected by with a joystick. Each trajectory consists of a pair of actions and corresponding observation frames from the agent's point of view.

\begin{figure}[h!]
\centering
        \begin{subfigure}[b]{0.43\columnwidth}
               \centering
    \includegraphics[width=\textwidth]{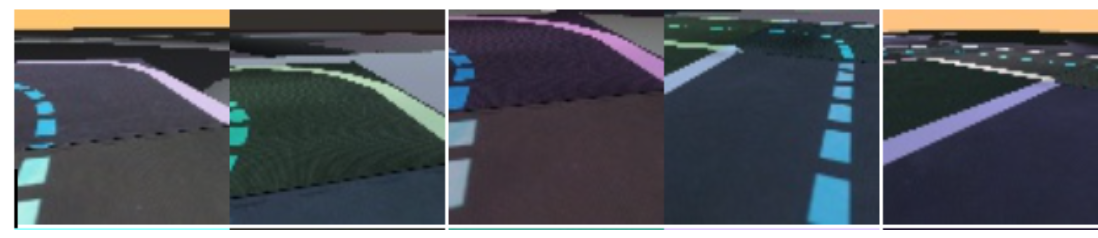}
    \caption{Simulated Images}
    \label{fig:sim_images}
        \end{subfigure}% \\
        \begin{subfigure}[b]{0.57\columnwidth}
               \centering
        \includegraphics[width=\textwidth]{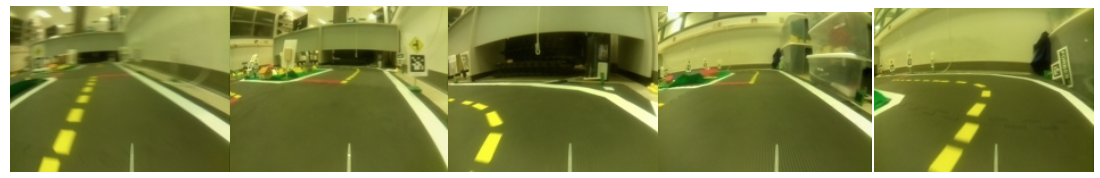}
    \caption{Real Images}
    \label{fig:real_images}
        \end{subfigure}%
\caption{Simulated and real images for the adversarial transfer training. Simulated images were collected from the Duckietown simulator. Real images were collected from the real Duckietown environment.}
\label{fig:sim_real_images}
\end{figure}

The sim/real frame dataset contains a list of image-label pairs, where the label corresponds to the domain (either sim or real). The images from simulator were collected using basic domain randomization with respect to camera height, angle, field of view, floor color, horizon color and pose of the robot.
% \begin{itemize}
%   \item camera height (distance from the ground to the camera): random in a range from 9.936cm to 11.664cm ($\pm8\%$);
%   \item camera angle: random in a range from 16º to 24º ($\pm20\%$);
%   \item camera field of view (FoV): random changes based on a 41.41-degree vertical FoV ($\pm20\%$);
%   \item distance between the robot's wheels: random in a range from 9.18cm to 11.22cm ($\pm10\%$);
%   \item floor color: random changes of the greyish color rgb(38, 38, 38) ($\pm30\%$);
%   \item horizon color: randomly selected from a blue sky color rgb(115, 209, 255) ($\pm10\%$), wall color rgb(163, 181, 71) ($\pm10\%$), greyish color rgb(38, 38, 38) ($\pm40\%$) and whitish color rgb (230, 230, 230) ($\pm40\%$);
%   \item position of the artificial sun: random in the simulation environment;
%   \item pose of the robot: randomly located on the road with random facing angle.
% \end{itemize}
% The randomization above are all uniformly distributed. The camera height, angle, FoV and robot's wheel distance are designed manually in order to match the variation of the duckiebots in the real world. The floor color, horizon color and lighting effects from the artificial sun simulate the actual driving scene in the real world, which makes sure the dataset contain scenes with various lighting conditions.
 The real images shown in Figure~\ref{fig:sim_real_images} were collected though the front camera of a physical Duckiebot by ensuring capture of different facing angles and positions on the road. 
%  The frames were extracted from the log files and scaled to $84 \times 84 \times 3$.

\subsection{Training}
\label{sec:training}
For all experiments, we train the model in a curriculum learning style during the pre-training (in sim) and fine-tuning (in real) phases. In practice, this means that while sampling trajectories for each batch, we consider those with shorter horizon lengths before the longer ones and the lane-following trajectories before the turning ones.
% Expert trajectories corresponding to a variety of different initial positions as described in Section~\ref{} are used for pre-training in simulation. After Adversarial Transfer, fine-tuning is done using expert trajectories in real, which are of lesser diversity with respect to initial poses (Section~\ref{}). We train the model once by using expert trajectories for both lane following and left turn, while subsequently performing tests in both the experiments. 

% In simulation, the training proceeds in a curriculum learning fashion, with straight road trajectories being used first followed by left turn trajectories and finally trajectories from a large map of the Duckietown environment.

\section{RESULTS}
% We first evaluated the performance of the proposed architecture in the simulation setup. Then we assessed the efficacy of the adversarial transfer and the performance of the architecture in the real setup. 
The performance of the framework has been measured by four metrics:  \textbf{outer loss} ($\mathcal{L}_{imitate}$), \textbf{inner loss} ($\mathcal{L}_{plan}$), \textbf{average reward per  time step} (simulation only), and \textbf{average completion rate} (fraction of the total distance to goal travelled by the Duckiebot before falling off the road averaged over all test instances with the same initial conditions). 
The reward function is given by
    \[
      r=\begin{cases}
                   v\cdot dir-10|d_c|,& \text{if on the right lane}\\
                   0,&\text{otherwise}\\
                \end{cases}
    \]
    where $v$ is the velocity of the Duckiebot, $dir$ is the moving direction of the Duckiebot and $d_c$ is the distance of the Duckiebot away from the right lane center. 
    %The average reward for single time step is then :    $    r_{avg} = \frac{r_{total}}{n}    $
    %where $n$ is the total number of time steps;

% \begin{table}[h!]
% \caption{Description of baselines and their corresponding indices.}
% \begin{tabular}{@{}cc@{}}
% \toprule
% Model Index &      Model Description                    \\ \midrule
% A &Final Model                  \\
% B& Without Stochasticity        \\
% C& Huber Loss Instead of MLP      \\
% D& Without Regularization        \\
% E& Without Regularization and Stochasticity    \\
% F& Huber Loss Instead of MLP and Without Regularization    \\
% G& Huber Loss Instead of MLP  and Without Stochasticity   \\
% H& Vanilla     \\ \bottomrule 
% \end{tabular}
% \end{table}

\subsection{Convergence Analysis of the Planner Module}

\begin{figure}[t]
        \begin{subfigure}[b]{0.5\columnwidth}
               \centering
    \includegraphics[width=\textwidth]{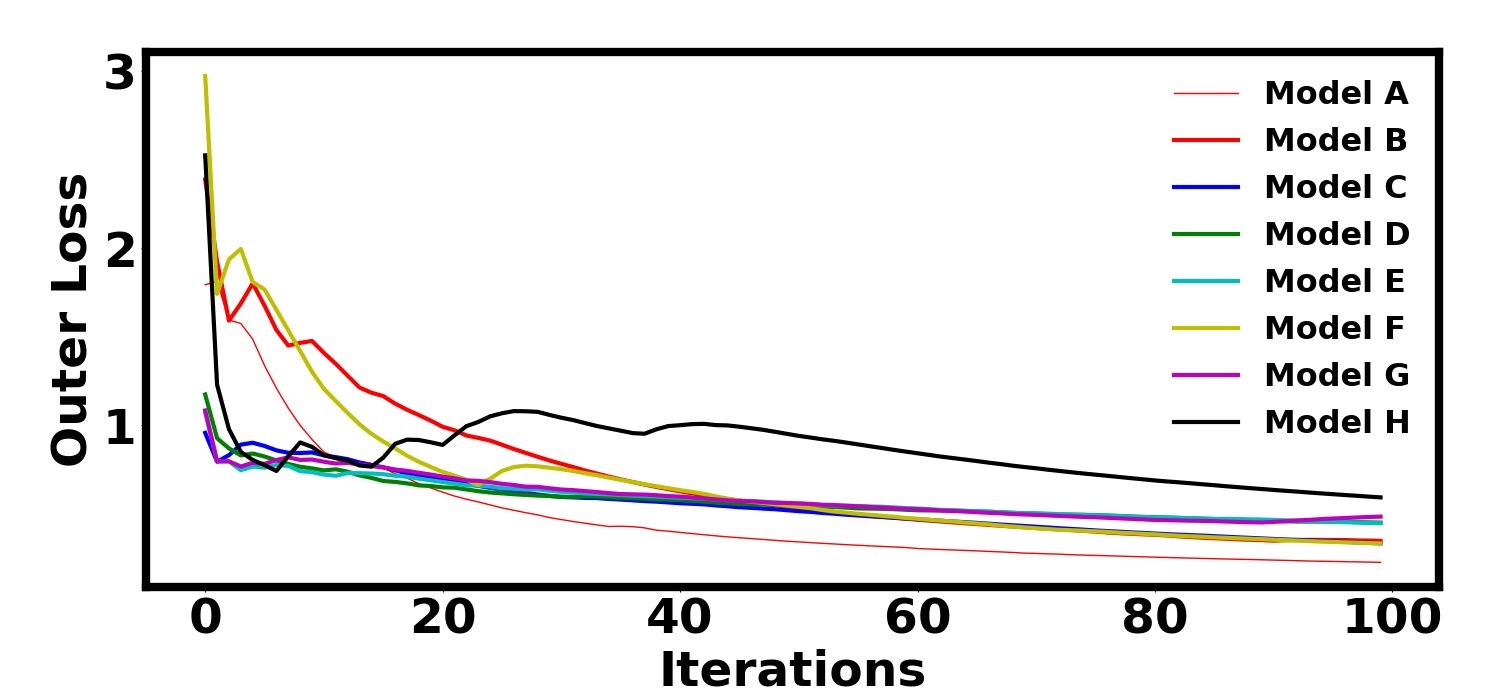}
    \caption{Model convergence on simulator data}
    \label{fig:simouterloss}
        \end{subfigure}%
        \begin{subfigure}[b]{0.5\columnwidth}
               \centering
   \includegraphics[width=\textwidth]{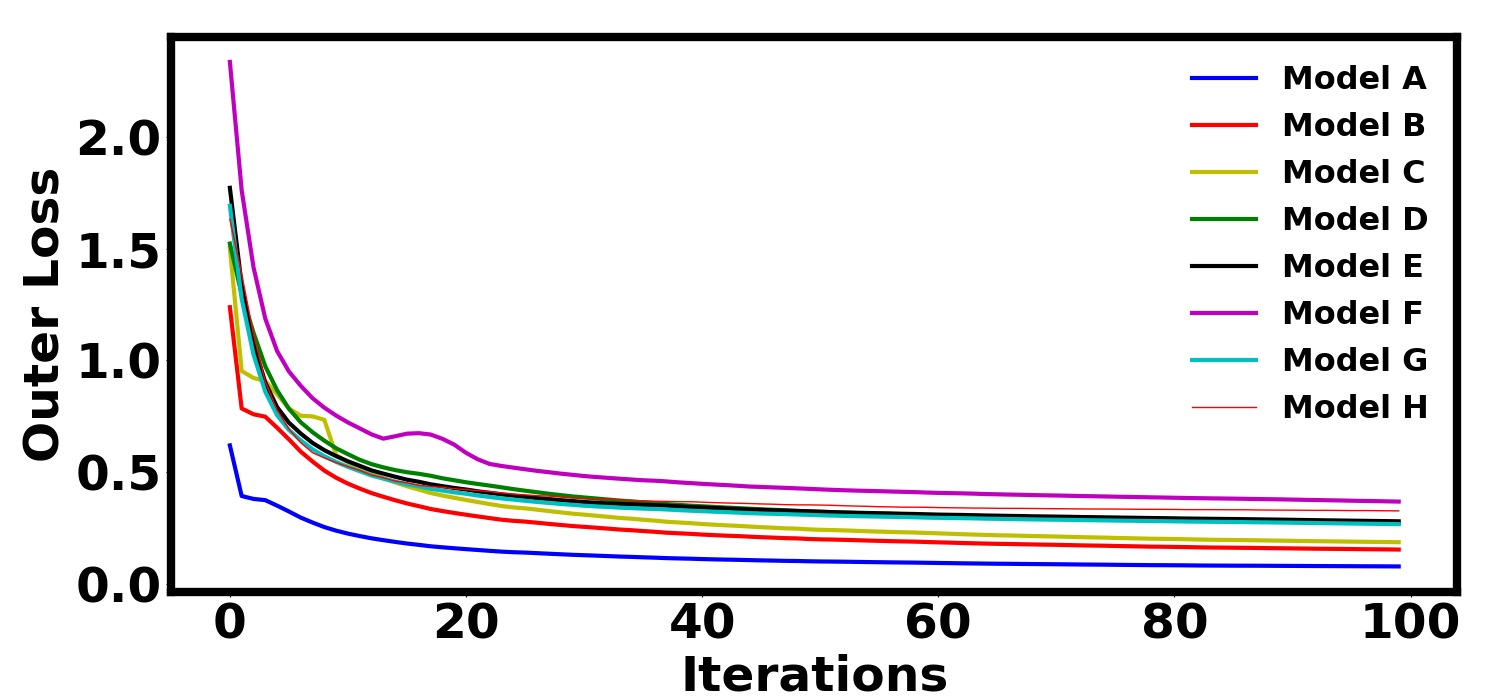}
    \caption{Model convergence on real data}
    \label{fig:realouterloss}
        \end{subfigure}%
 \caption{Evaluation of various baselines models on Duckietown simulation (a) and real (b) environment showing the convergence of outer loss as training progresses.  (Model A denotes our final planner, B is the version without Stochasticity, C is the version with Huber Loss instead of MLP as the planning loss, D does not incorporate regularization, E is D sans Stochasticity, F is C sans Regularization, G is C sans Stochasticity and H is the vanilla UPN~\cite{UPN} planner). }   
\end{figure}

% \begin{figure}
% \centering
% \includegraphics[width=\columnwidth]{LeftTurnCompare.png}
% \caption{Evaluation of various baselines models on Duckietown Simulation environment. The Outer Loss is the imitation learning loss on  test trajectories. The graph shows the convergence of outer loss for each model as training progresses. Less is Better}
% \label{fig:simouterloss}
% \end{figure}

% \begin{figure}
% \centering
% \includegraphics[width=\columnwidth]{real-outerloss.png}
% \caption{Evaluation of various baselines models on Duckietown Real environment via sim-to-real transfer. The Outer Loss is the imitation learning loss on  test trajectories. The graph shows the convergence of outer loss for each model as training progresses. Less is Better}
% \label{fig:realouterloss}
% \end{figure}

Here we analyze the efficacy of the key components of the planner module proposed in Sec.~\ref{sec:method}. Fig.~\ref{fig:simouterloss} depicts the convergence of the models during pre-training in simulation through the training procedure in Alg.~\ref{alg:pre-training}. Fig.~\ref{fig:realouterloss} shows the convergence of the models during fine-tuning by the use of real expert trajectories. It is evidenced from both the figures that Model A, which is our final model incorporating all the components described in Sec.~\ref{sec:method} has a much steeper convergence rate and also converges to a better optimum.

\subsection{Evaluation on Duckietown Simulation Environment}

\begin{figure}[t]
        \begin{subfigure}[b]{0.51\columnwidth}
               \centering
    \includegraphics[width=\textwidth]{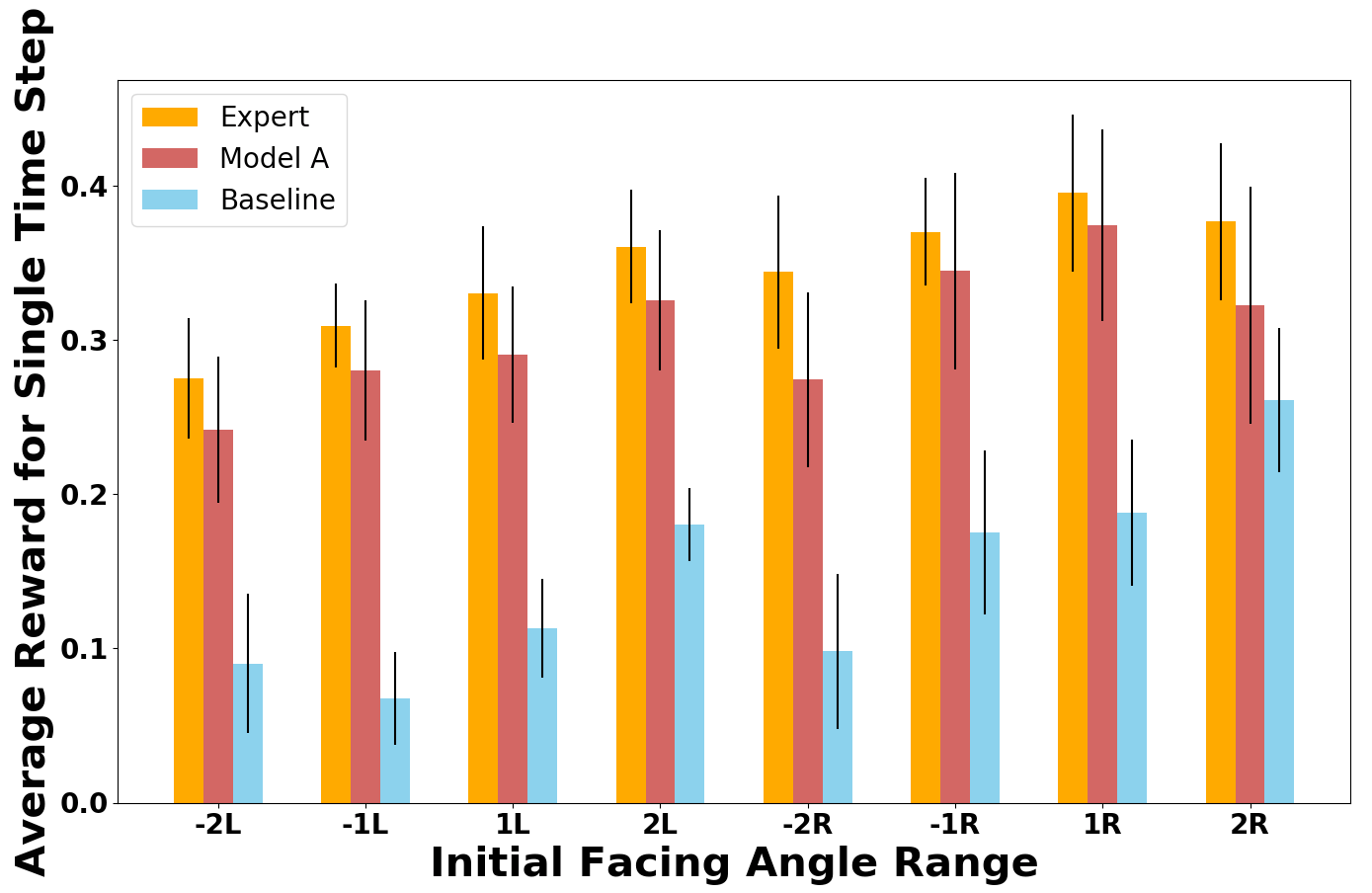}
    \caption{Average Reward (LF)}
    \label{fig:avg_r_lf}
        \end{subfigure}% \\
        \begin{subfigure}[b]{0.49\columnwidth}
               \centering
        \includegraphics[width=\textwidth]{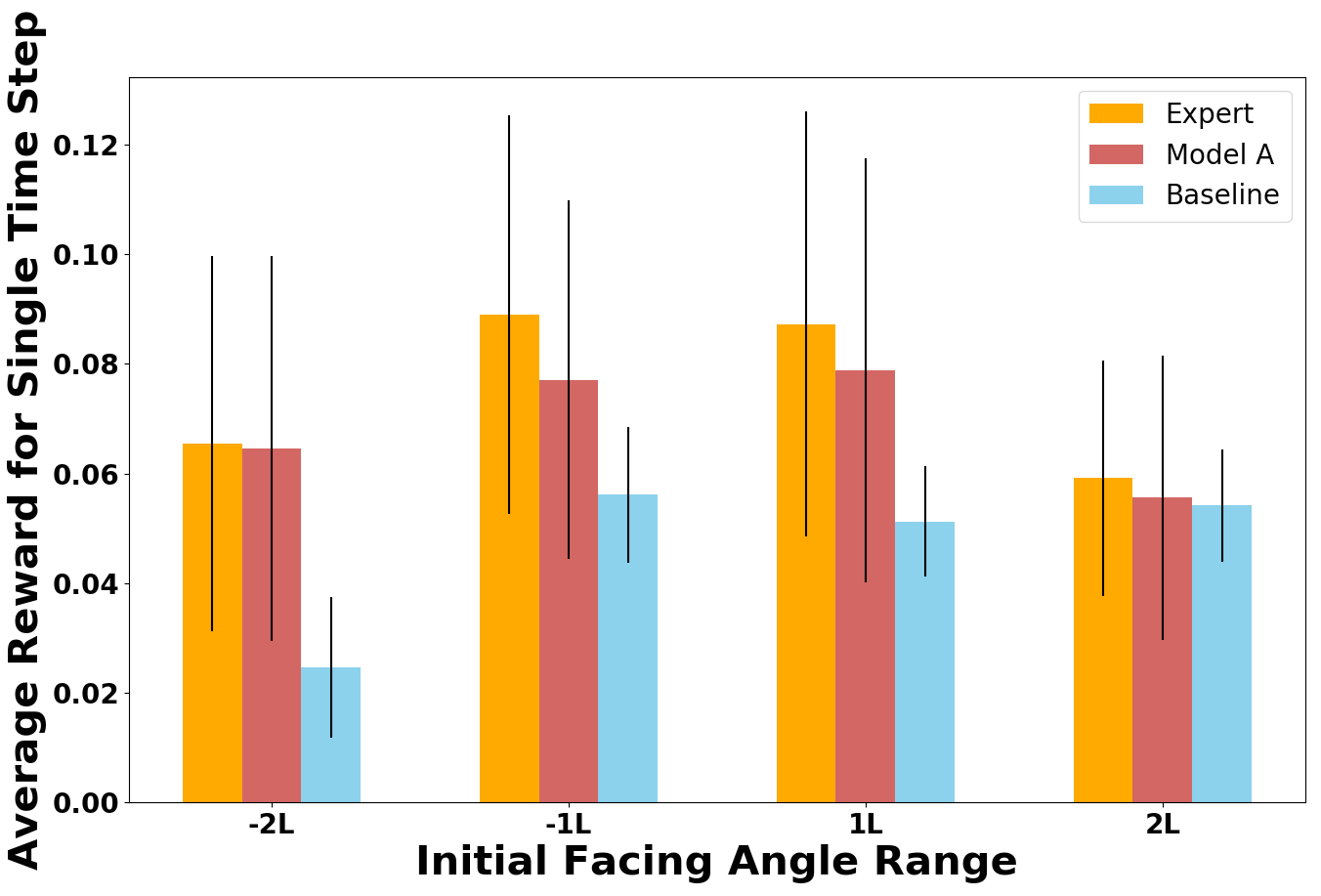}
    \caption{Average Reward (LT)}
    \label{fig:avg_r_lt}
        \end{subfigure}%
    \quad
            \begin{subfigure}[b]{0.525\columnwidth}
               \centering
        \includegraphics[width=\textwidth]{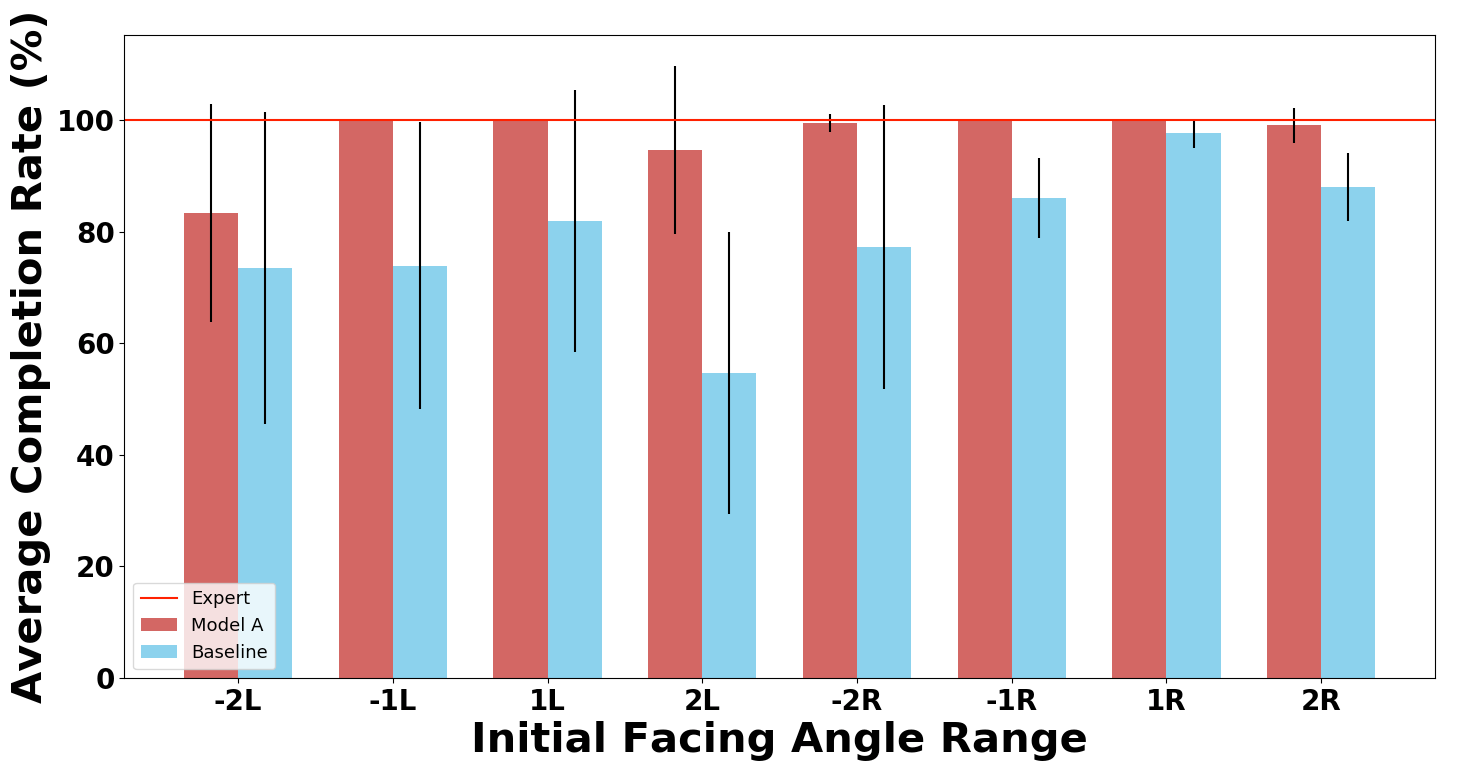}
    \caption{Completion Rate (LF)}
    \label{fig:avg_com_lf}
        \end{subfigure}%
        \begin{subfigure}[b]{0.475\columnwidth}
               \centering
        \includegraphics[width=\textwidth]{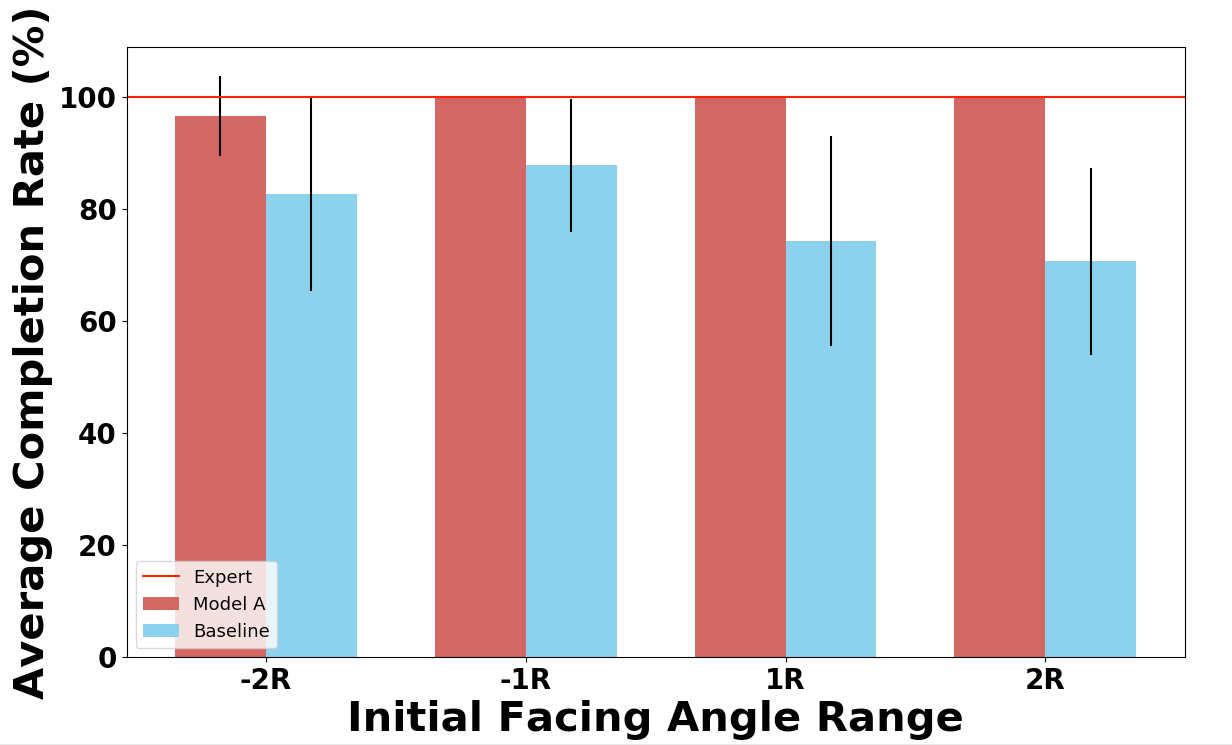}
    \caption{Completion Rate (LT)}
    \label{fig:avg_com_lt}
        \end{subfigure}%
 \caption{Evaluation of the average time step reward and average completion rate on Duckietown Simulator (Notation: L - left lane, R - right lane; $-2$ - (-30º, -15º), $-1$ - (-15º, 0º), $1$ - (0º, 15º), $2$ - (15º, 30º); LF - lane following; LT - left turn).}  
 \label{fig:sim_results}
\end{figure}

% \begin{figure}[t]
% \centering
% \includegraphics[width=\columnwidth]{average_time_step_reward.png}
% \caption{Evaluation of the average time step reward on Duckietown Simulator. The horizontal axis is the testing horizon. The vertical axis is the initial facing angle range. Each grid (horizon-angle pair) was tested by 10 testing samples. }
% \label{fig:innerloss}
% \end{figure}

% \begin{figure}[h!]
% \centering
% \includegraphics[width=\columnwidth]{average_time_step_reward_left_turn.png}
% \caption{}
% \label{fig:innerloss}
% \end{figure}

% Here we analyze the performance of different variants of the proposed approach and compare their performance with the final proposed model.  We Evaluated 

% As shown in Fig.~\ref{fig:simouterloss}, the convergence of Model A is the fastest, while that of Model H is erratic. 

We now evaluate the performance of our model after pre-training in simulation through the training procedure described in Sec.~\ref{sec:training}. The results of the lane following test are shown in Fig.~\ref{fig:avg_r_lf} and Fig.~\ref{fig:avg_com_lf} and that of the left turn test are highlighted in Fig.~\ref{fig:avg_r_lt} and Fig.~\ref{fig:avg_com_lt}. We observe that Model A significantly outperforms the baseline UPN model. We claim that this improvement in simulation is a crucial stepping stone for effective sim-to-real transfer.

\subsection{Evaluation of the Inner-Loop Loss Function}

\begin{figure}[tb]
%\begin{minipage}[c]{0.69\columnwidth}
\includegraphics[width=\columnwidth]{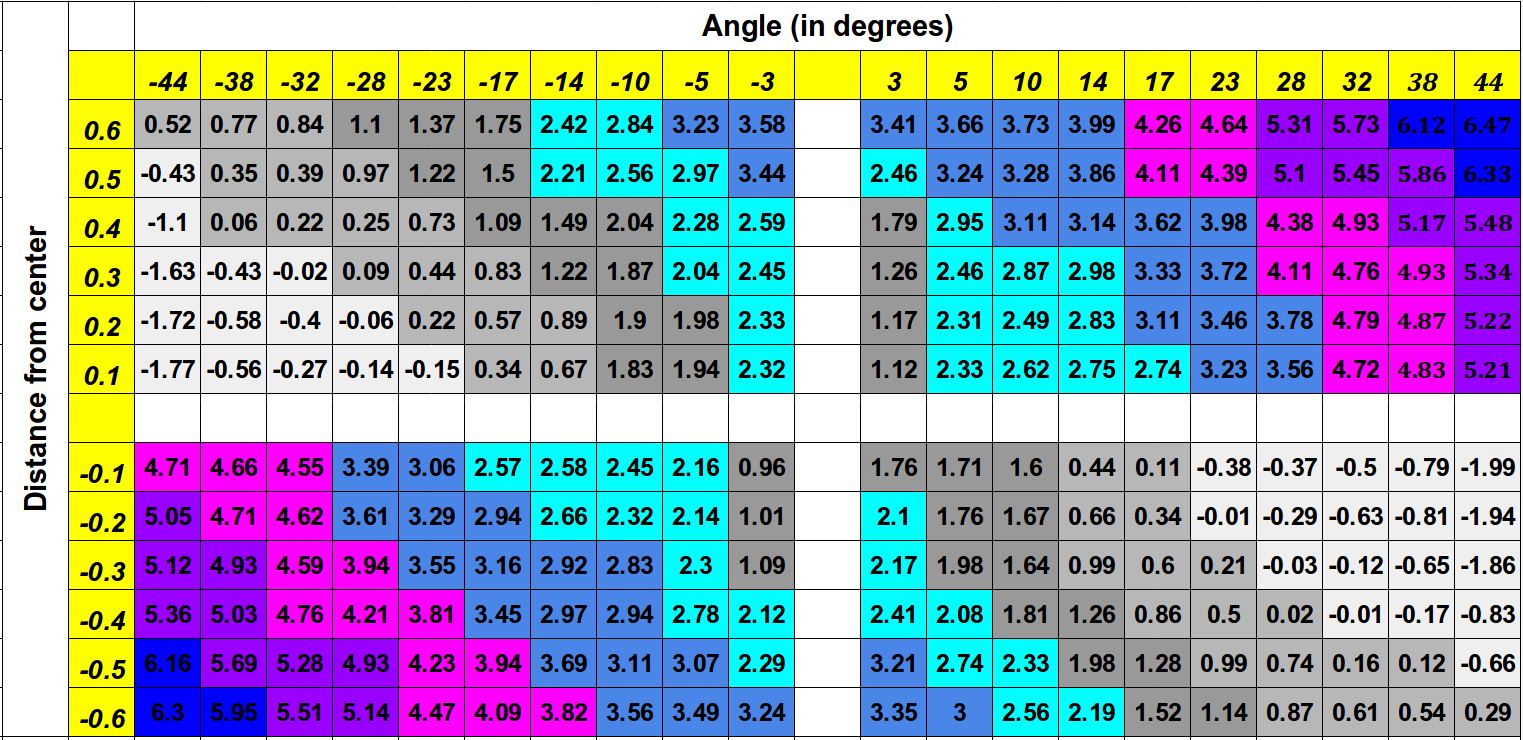}
%\end{minipage}\hfill
%\begin{minipage}[c]{0.3\columnwidth}
%\vspace{0.1cm}
\caption{Evaluation of the inner loop loss function on Duckietown Simulator. The evaluation is of Model A that has been trained on horizons of different lengths from 30 to 300 via curriculum learning.}
\label{fig:innerloss}
%\end{minipage}
\end{figure}

In our planner, we have a MLP as the inner-loop loss function whose parameters are learned in the outer imitation learning loop as described in Sec.~\ref{sec:MLP}. 
%Since it is not evident that this loss function is indeed suitable (unlike a squared loss error where the intuition of minimization is clear), we do an analysis of judging the efficacy of its suitability. 
After training the model, we fix the parameters of the MLP inner-loss and test for its value in different positions on the road. Intuitively, the value of the loss inferred by this function should he high near the center of the lane and should increase away from the center. Empirical evaluations in Fig.~\ref{fig:innerloss} justify that the loss function conforms to our intuition about its desired behavior.

% \begin{figure}
% \centering
% \includegraphics[scale=0.4]{ad_trasfer_loss.png}
% \caption{ Evaluation of the Adversarial transfer part of the framework by fixing the number of simulation trajectories used in pre-training at 2000 and the number of real trajectories used in fine-tuning to 100. The plot shows the variation of outer loss. Less is better. }
% \label{fig:ad_transfer_loss}
% \end{figure}
\begin{figure}
%\begin{minipage}[c]{0.4\columnwidth}
\includegraphics[width=\columnwidth]{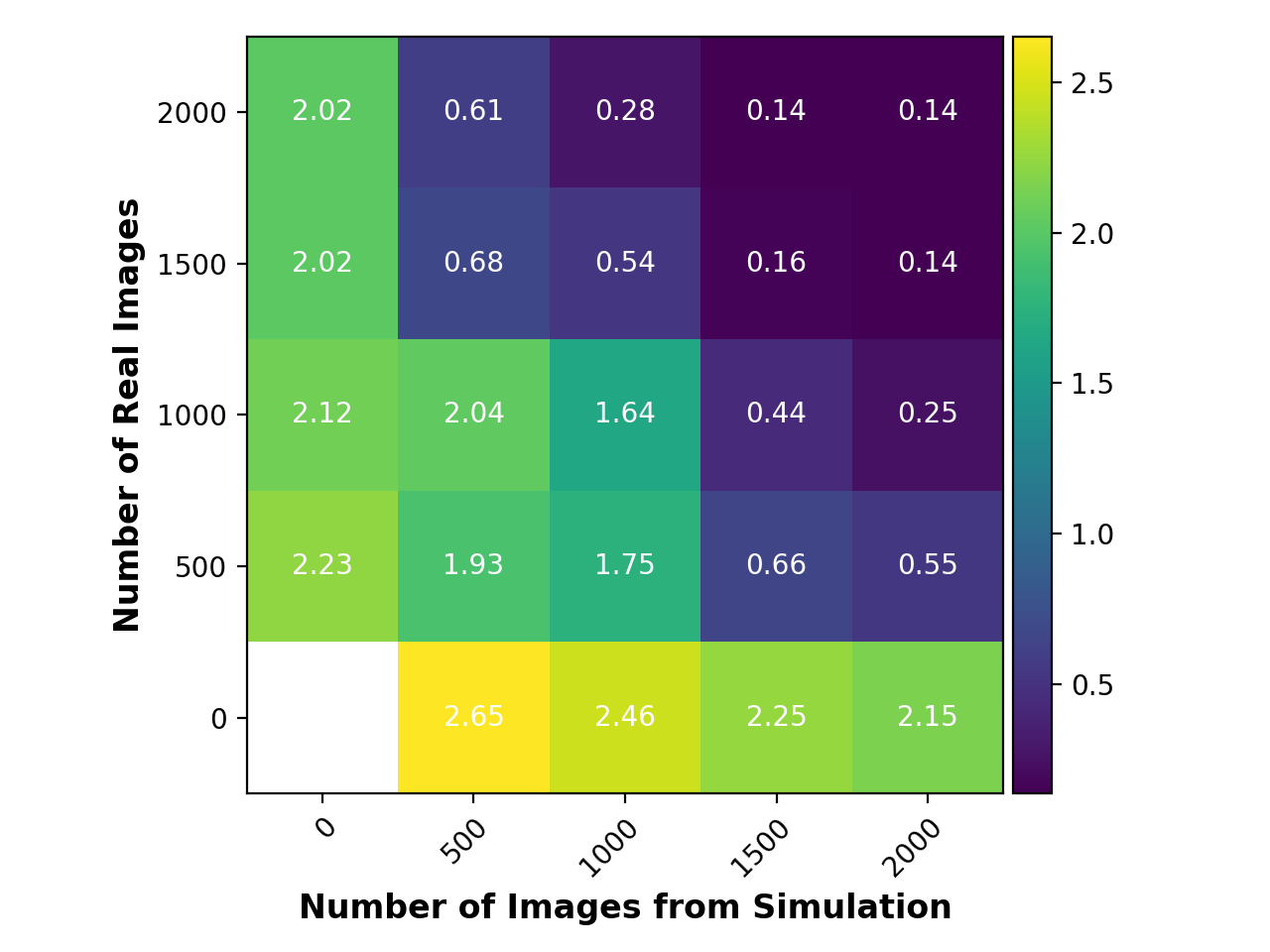}
%\end{minipage}\hfill
%\begin{minipage}[c]{0.6\columnwidth}
\caption{Evaluation of the performance of adversarial transfer in terms of loss as a function of the number of real and simulated images used in training, after pre-training in simulation with 2000 expert demos. }\label{fig:ad_transfer_loss}
%\end{minipage}
\end{figure}

%\begin{figure*}[tb]
%        \begin{subfigure}[b]{0.24\textwidth}
%               \centering
%    \includegraphics[width=\textwidth]{ad_trasfer_loss.png}
%    \caption{Adversarial Transfer}
%    \label{fig:ad_transfer_loss}
%        \end{subfigure}
%        \begin{subfigure}[b]{0.37\textwidth}
%               \centering
%        \includegraphics[width=\textwidth]{real_result_lt.png}
%    \caption{Lane Following}
%    \label{fig:real_lt}
%        \end{subfigure}%
%         \begin{subfigure}[b]{0.37\textwidth}
%               \centering
%        \includegraphics[width=\textwidth]{real_result_sr_1.png}
%    \caption{Left Turn}
%    \label{fig:real_lf}
%        \end{subfigure}%
% \caption{(a) Evaluation of the Adversarial transfer part. The plot shows %the variation of outer loss. Less is better. \textbf{Notation:} L - left %lane center, M - Middle line center, R - right lane center; (b) Average %completion rate on the left turn test (real)(b) Average completion rate %on the lane following test (real)}  
% \label{fig:sim_results}
%\end{figure*}

\subsection{Efficacy of the Transfer to the Real Robot}

\begin{figure*}[t!]
\begin{subfigure}{0.58\textwidth}
\centering
\includegraphics[width=\textwidth]{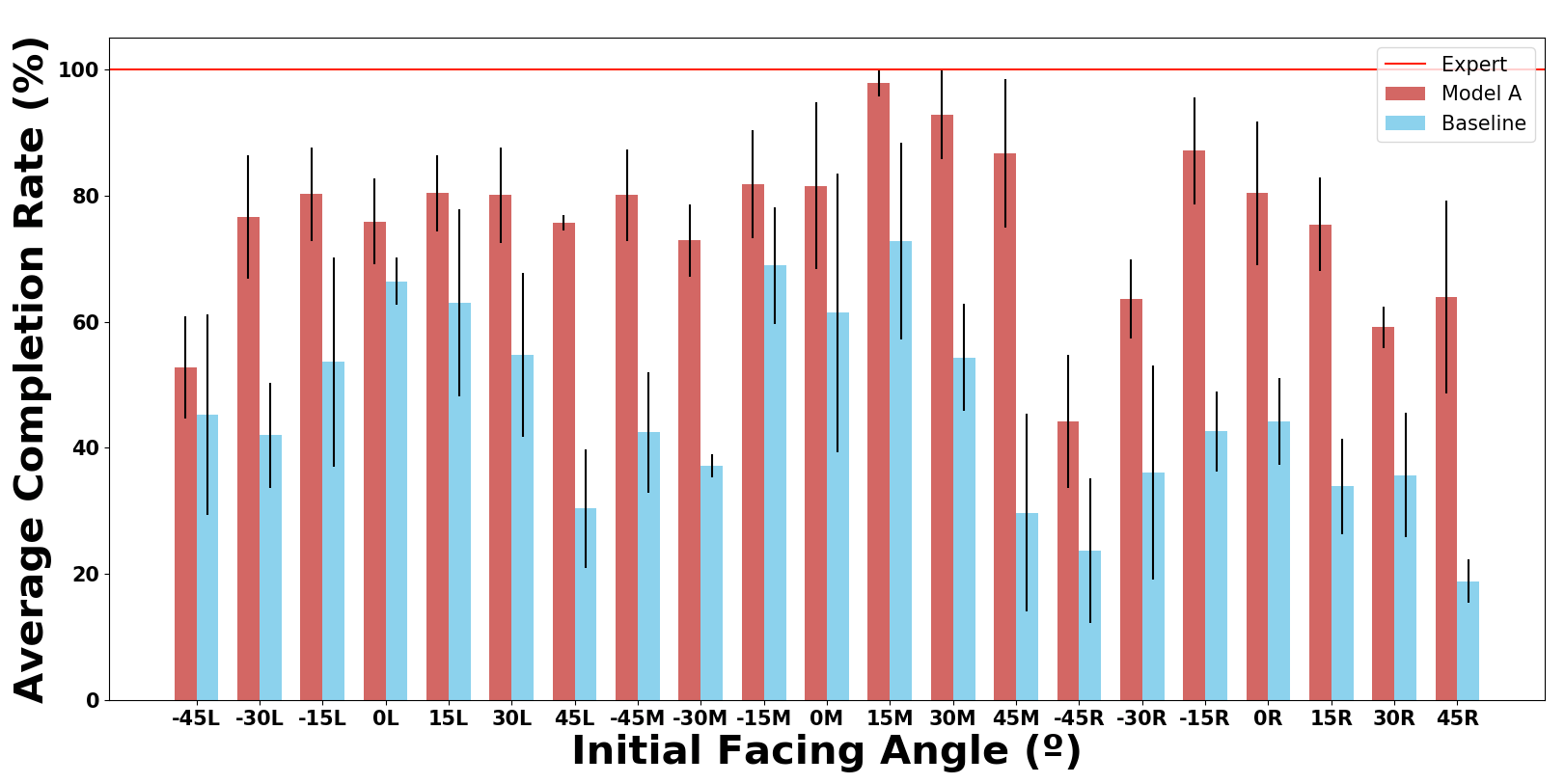}
\caption{ Average completion rate on the lane following test (real).}
\label{fig:real_lf}
\end{subfigure}
\begin{subfigure}{0.38\textwidth}
\centering
\includegraphics[width=\textwidth]{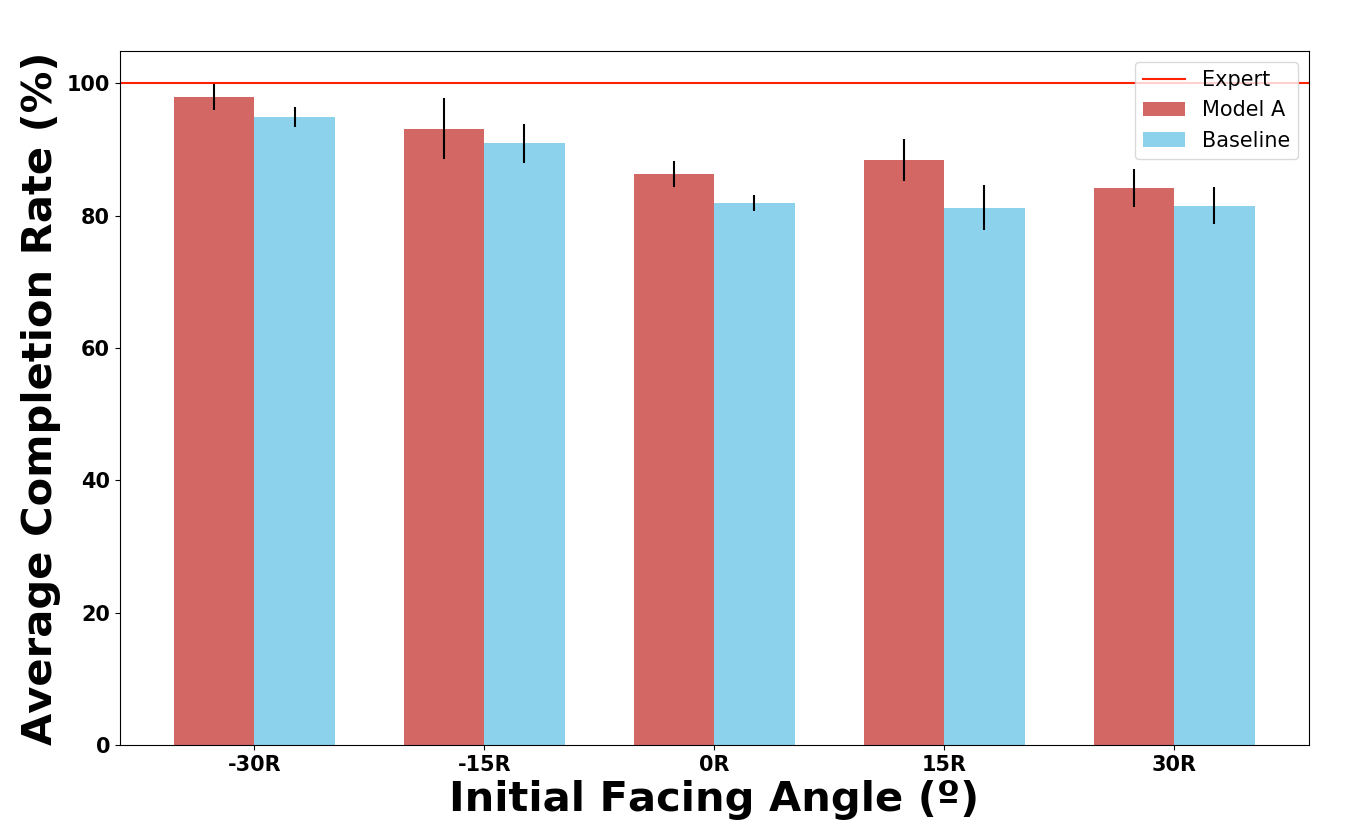}
\caption{Average completion rate on the left turn test (real).}
\label{fig:real_lt}
\end{subfigure}
\caption{Performance of our method on the real robot (Notation: L - left lane center, M - Middle line center, R - right lane center;)}
\end{figure*}

%This section presents the core of our results on sim-to-real transfer. 
After pre-training in simulation and performing adversarial domain transfer, we fine-tune the model in the real setup. The architecture used is our final Model A. The results of the lane following test are shown in Fig.~\ref{fig:real_lf} and that of the left turn test are highlighted in Fig.~\ref{fig:real_lt}. We use domain randomization~\cite{dynamicsRand} as baseline against which we compare our sim-to-real transfer architecture\footnote{For a video of the real robot results please refer to \href{https://www.youtube.com/watch?v=KhSUWapOgXg}{this link}}. 

It is interesting to note that our model performs quite well ($>50\%$ average completion rate) even for the most difficult case of navigation starting from the center of the left lane with an initial facing angle of -45º. Also of note is the fact that the performance on left-turn is quite good for our model. This is indicative of the curriculum learning framework, which first learns lane following followed by turning (in training) yielding noticeable gains during testing. We also evaluated how many real and simualated images were required for convergence of the adversarial loss, with results presented in Fig.~\ref{fig:ad_transfer_loss}, and also how many real trajectories were needed to achieve an equivalent outer-loop loss with and without our transfer learning pipeline, with results presented in Table~\ref{table:data_efficiency}. From these two results, we see that our method preferentially uses ``off-policy'' data to save the amount of on-policy expert trajectories needed on the real robot. %Our method requires very few expert trajectories on the real physical robot. 
%convSince in our model, the latent state is not enforced to explicitly correspond to position (unlike ~\cite{ADT}),  we cannot evaluate the efficacy of the adversarial transfer directly. Instead, we fix the number of trajectories used for pre-training in sim (to 2000) and fine-tuning in real (to 100), and vary the number of images in sim and real during training.  shows the variation of the outer loss with different sim-real image combinations. 
%Interestingly, we observe that the benefit of having beyond  1500 is not profound.

% \begin{table}[h!]
% \caption{Description of baselines and their corresponding indices.}
% \begin{tabular}{@{}ccc@{}}
% \toprule
% Outer loss &     $#$Real Trjs (Direct) & $#$Real Trjs (Transfer)                    \\ \midrule
% 0.10 &   1250 &     230        \\
% 0.15&  1150   &   180  \\
% 0.20& 950  &  120  \\
% 0.25& 750   &  75   \\
% 0.30 & 500  & 50 \\
% 0.35 & 200 & 25 \\\bottomrule 
% \end{tabular}
% \end{table}

\vspace*{-0.15cm}
\begin{table}[h!]
\centering
\caption{The number of real trajectories required in the proposed sim-to-real transfer compared to training the model directly  without sim-to-real. }
\label{table:data_efficiency}
\begin{tabular}{@{}ccccccc@{}}
\toprule
Outer loss &  0.10    &0.15 & 0.20 & 0.25 & 0.30 & 0.35                   \\ \midrule
No. of Real Trjs (Direct) &   1250 &   1150 & 950 & 750 & 500 & 200        \\ \midrule
No. of Real Trjs (Transfer)   &   230 & 180 & 120 & 75 & 50 & 25\\ \bottomrule 
\end{tabular}
\end{table}

\vspace{-0.4cm}
\section{CONCLUSION}
We present a framework for gradient-based planning and transfer from sim-to-real. We demonstrated through experimentation that the proposed method achieves significant performance gains in the real environment by learning a robust policy in simulation followed by a successful adversarial transfer.
% \section*{ACKNOWLEDGMENT}
% Mila GPUs et al.
%\clearpage

\bibliographystyle{IEEEtran}
\bibliography{icra}

\end{document}